\documentclass[conference]{IEEEtran}
\IEEEoverridecommandlockouts
\usepackage{cite}
\usepackage{amsmath,amssymb,amsfonts,bm}
\usepackage{algorithmic}
\usepackage{graphicx}
\usepackage{textcomp}
\usepackage{xcolor}
\usepackage{enumitem}
\usepackage{comment}
\usepackage{times}
\usepackage{epsfig}
\usepackage{multirow}
\usepackage{booktabs}
\usepackage{subfig}
\usepackage[pagebackref,breaklinks,colorlinks]{hyperref}
\def\BibTeX{{\rm B\kern-.05em{\sc i\kern-.025em b}\kern-.08em
    T\kern-.1667em\lower.7ex\hbox{E}\kern-.125emX}}
\begin{document}

\title{Measuring Bias in AI Models: \\ An Statistical Approach Introducing N-Sigma}


\author{Daniel DeAlcala, Ignacio Serna, Aythami Morales, Julian Fierrez, Javier Ortega-Garcia\\
Biometrics and Data Pattern Analytics Lab, Universidad Autonoma de Madrid, Spain\\}


\maketitle

\begin{abstract}
The new regulatory framework proposal on Artificial Intelligence (AI) published by the European Commission establishes a new risk-based legal approach. The proposal highlights the need to develop adequate risk assessments for the different uses of AI. This risk assessment should address, among others, the detection and mitigation of bias in AI. In this work we analyze statistical approaches to measure biases in automatic decision-making systems. We focus our experiments in face recognition technologies. We propose a novel way to measure the biases in machine learning models using a statistical approach based on the N-Sigma method. N-Sigma is a popular statistical approach used to validate hypotheses in general science such as physics and social areas and its application to machine learning is yet unexplored. In this work we study how to apply this methodology to develop new risk assessment frameworks based on bias analysis and we discuss the main advantages and drawbacks with respect to other popular statistical tests.   
\end{abstract}

\begin{IEEEkeywords}
Artificial Intelligence, AI, Bias, Explainable, Risk Assessment, Trustworthiness, 5-Sigma
\end{IEEEkeywords}

\section{Introduction}

Artificial Intelligence (AI) can play an important role to achieve the Sustainable Development Goals (SDGs) by 2030 \cite{UN21}.  AI brings enormous benefits in several critical areas for our society (e.g., health, security, sustainability), but it can also significantly compromise the safety of citizens worldwide. The development of a Responsible AI technology needs an international multidisciplinary effort to ensure the trustworthiness, sustainability, and safety. This effort involves a multi-stakeholder work including academia, industry, civil society, and public agencies, among others.  

\begin{figure}[t!]
\centering
\includegraphics[width=1\columnwidth]{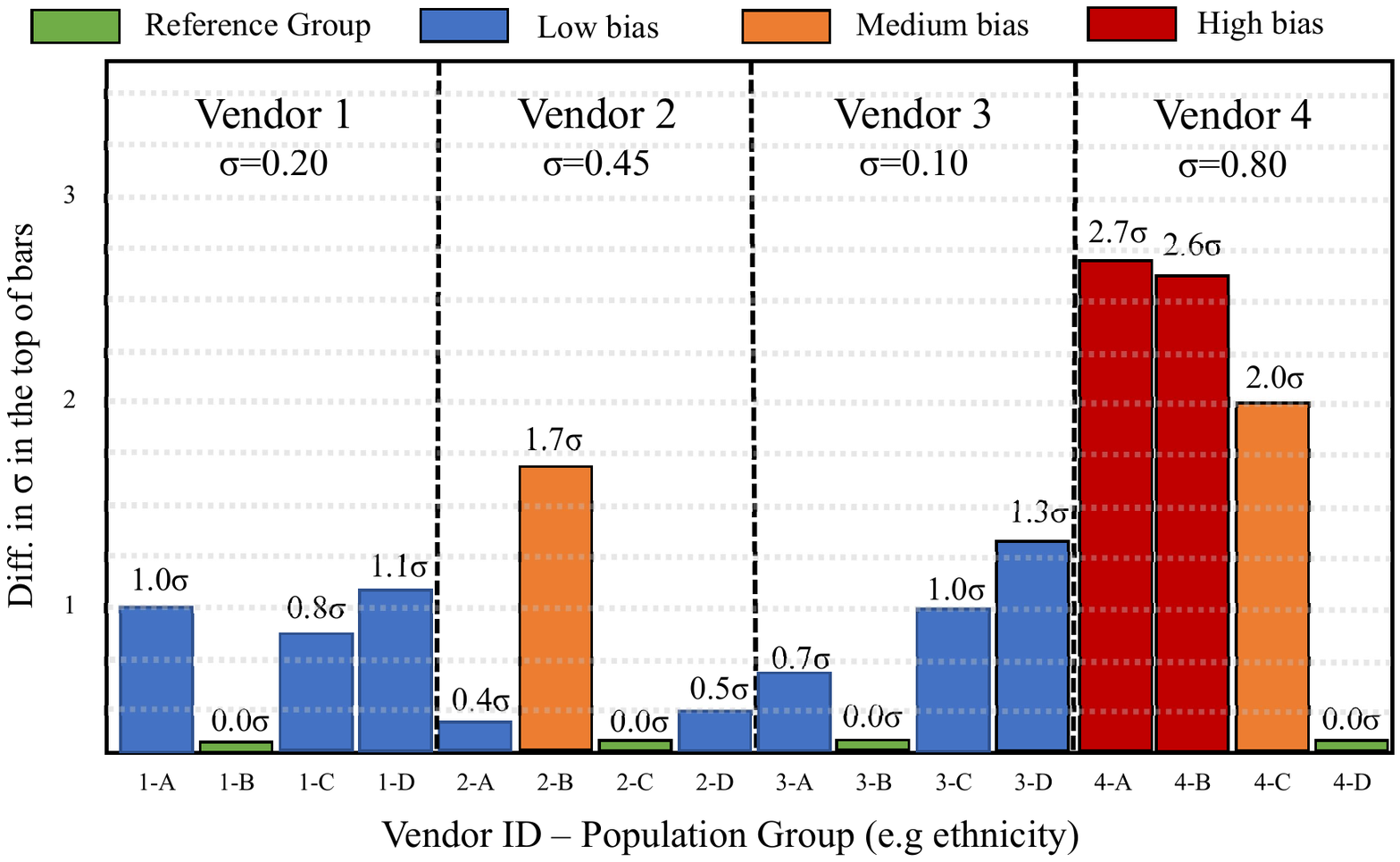} 
\vspace{-8mm}
\caption{Performance differences of 4 AI models (e.g., Face Recognition technology) evaluated over 4 different demographic groups (A,B,C,D). The difference is measured in number of sigmas ($\sigma$) with respect to a reference group. The bias level is represented with different colors. 
}
\label{fig:Nsigmas_idea}
\end{figure}

The absence of international standards for the development of Responsible Artificial Intelligence has motivated a wide variety of approaches \cite{lepri2021ethical, rahwan2019machine}. The regulation is moving from a technology-based framework to a risk-based framework \cite{EC20}. The new regulatory framework proposed by the European Union defines 4 levels of risk in AI: \textit{i)} Unacceptable, \textit{ii)} High, \textit{iii)} Limited, and \textit{iv)} Minimal. As an example, high-risk technologies will be subject to strict obligations including adequate risk assessment and mitigation systems. This risk-based framework requires protocols and technologies capable of assessing and explaining the results of AI systems based on parameters beyond the traditional performance metrics (e.g., overall accuracy).     

How to measure or assess the fairness of an automatic-decision algorithm is not a trivial task. Fairness is a human concept that can be mathematically defined in different ways \cite{hardt2016equality}. Traditionally, fairness is measured as a difference in performance between population subgroups (e.g., performance for different demographic groups). The literature has proposed several approaches to measure such a difference with traditional statistical methods \cite{godbole2022demographic, vzliobaite2017measuring} or machine learning approaches \cite{hardt2016equality, mehrabi2021survey}, but none of these methods has been yet adopted as a widely recognized standard. 

On the other hand, the $5$-sigma approach is widely used for statistical analyses in many fields including natural \cite{lyons2013sigma} and human sciences \cite{leonard2003schaum}. In this work we extend this $5$-Sigma approach and apply it to bias analysis of data-driven learning models (see Fig. \ref{fig:Nsigmas_idea}). The contributions of this work are:
\begin{itemize}
    \item The proposal of a common experimental protocol to achieve a fairer and more standardized evaluation of AI models. 
    \item We analyze two pointwise metrics and a traditional distribution metric for bias analysis in machine learning models. (More specifically, in discrimination-aware face recognition models.) 
    \item We extend 5-Sigma into $N$-Sigma for bias assessment of machine learning. This proposed extension is compatible with a risk-based evaluation framework where a variable ($N$) can be associated to each risk level (see Fig. \ref{fig:Nsigmas_idea}). 
     
\end{itemize}

\section{Related Works}

Bias in ML systems is an increasingly studied topic for which various notions of fairness have been applied \cite{verma2018fairness,mehrabi2021survey}. The most common way to measure bias is through performance in demographic groups, but it is not the only way. Researchers have also looked at how models respond at the level of activation and how this is different across different groups \cite{nagpal2019Face,serna2020insidebias}.

Among other AI application fields, face biometrics is perhaps the most popular and evolved one regarding bias analysis \cite{2021_TTS_Biases_Terhorst}. From the fairness criteria proposed in the literature, the statistical parity criterion is inadequate for Face Recognition (FR) models. The work \cite{locatello2019fairness} shows that a perfect model does not imply demographic parity with entangled variables, which is the case of FR, where sensitive demographic characteristics are linked to identity. It is unreasonable to think that the outcome of a face recognition system is independent of a person's ethnicity when a white user is trying to impersonate a black user. Equalized odds are often used; for example, the NIST report uses false negatives and false positives for each demographic group to measure the fairness \cite{grother2019FRVT}.

Except a few exceptions \cite{serna2022sensitiveloss}, most of the literature studying bias in facial algorithms does not clearly define what bias is and merely shows that the performance varies between population groups. Recent research is attempting to mitigate biases after quantifying them. These recent papers typically use a form of standard deviation of the algorithm performance across individuals of different populations as a measure for bias, both implicitly and explicitly \cite{gong2021gac,wang2019mitigate,2021ramface,serna2022sensitiveloss}.

\subsection{Bias in face recognition}
The number of academic studies analyzing the fairness of face recognition algorithms has grown significantly in recent years, and the number of published works pointing out the biases in the results of face detection \cite{buolamwini2018GenderShades} and recognition algorithms is large \cite{klare2012demographic,isabelle2019DemogPairs,drozdowski2020bias,2021_TTS_Biases_Terhorst}.
Facial recognition systems can suffer from a variety of biases, ranging from those arising from unconstrained environmental variables such as illumination, pose, expression, and face resolution, from systematic errors such as image quality \cite{2022_CSUR_FaceQsurvey_Torsten,2022_WACVw_FaceQvec_JHO}, and from demographic factors \cite{2018_TIFS_SoftWildAnno_Sosa} like age, sex, and race. Among these different covariates, the skin color is repetitively remarked as a factor with high impact in the performance \cite{chellapa2019face}.

\section{Database and Models}\label{sec:Models}

In our experiments we used  Racial Faces in the Wild (RFW) \cite{wang2019RFW}. This database is divided into four demographic classes: Caucasian, Indian, Asian, and African. Each class has about $10K$ images of $3K$ individuals. There are no major differences in pose, age, and sex distribution between Caucasian, Asian, and Indian groups. The African group has a smaller age difference than the others, and while females account for approximately 35\% in the other groups, they account for less than 10\% in the African group.

The model used is a ResNet-100 network \cite{he2016resnet}, trained on the MS1Mv3\footnote{\url{https://github.com/deepinsight/insightface/tree/master/recognition/_datasets_\#ms1m-retinaface}} database \cite{guo2016ms} (93K identitites and 5.2M images) with ArcFace \cite{deng2019arcface} loss function. A model with 101 convolutional layers and 44 million parameters.

When using facial recognition systems in verification mode, two faces are assigned the same identity if their similarity distance is smaller than a threshold $\tau$. The similarity is computed between the two face descriptors $\textbf{x}_r$ and $\textbf{x}_s$ obtained from a face model. A similarity score is known as a genuine score or authentic score if it is the result of matching two samples of the same biometric trait of a user. It is known as an impostor score if it involves comparing two biometric samples originating from different users \cite{ross2009biometrics}. Several metrics can be used to compute similarity, the two most frequent are euclidean distance and cosine similarity.

From the similarity results, EER (Equal Error Rate) and TPR (True Positive Rate) are computed for a specific threshold. The decision threshold $\tau$ for each model is different and is set using genuine and impostor comparisons. The EER is the error at a given threshold at which FMR (False Match Rate) and FNMR (False Non-match Rate) are equal. The TPR is the probability of correctly identifying two user samples as being from the same user.

\begin{figure*}[t]
\centering
\includegraphics[width=0.95\textwidth]{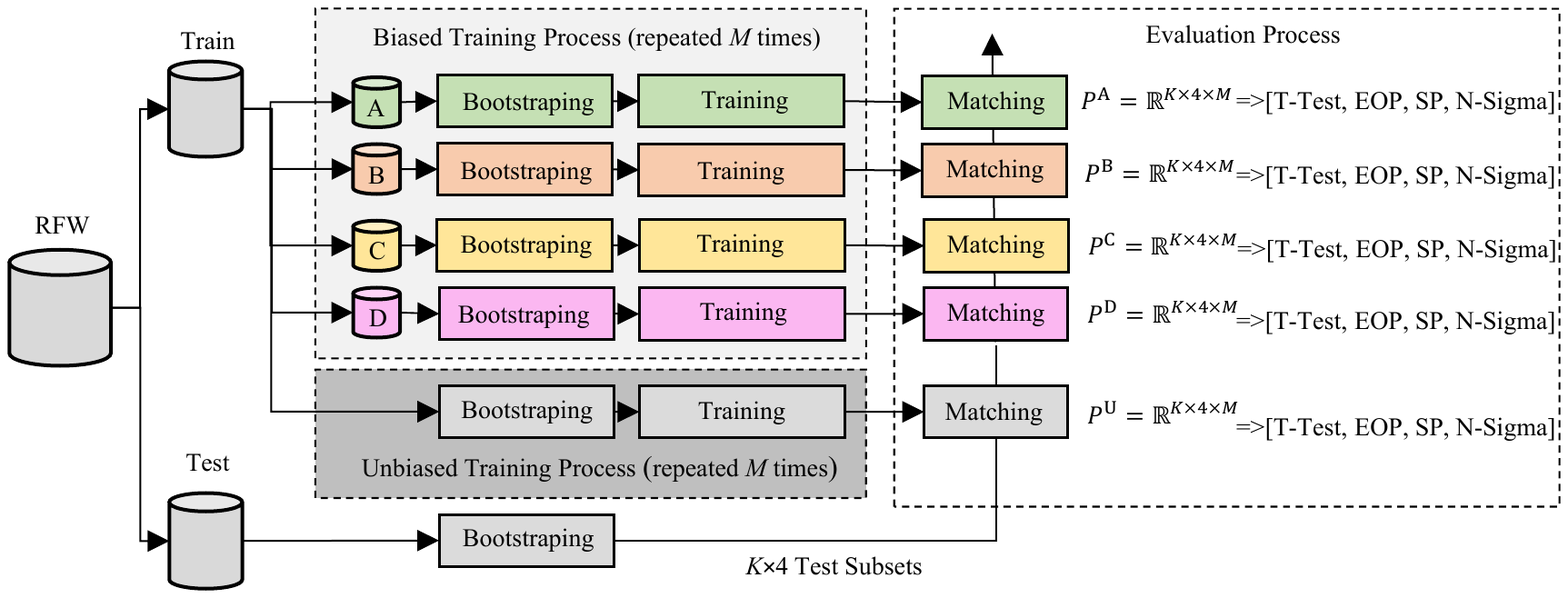} 
\caption{Experimental Framework for the analysis of biased learning processes. A, B, C, and D represent different demographic groups used to introduce bias in the learning process.}
\label{fig:Experimental_Framework}
\vspace{-4mm}
\end{figure*}

\section{Methods: Experimental Protocol} \label{section:Protocol}

\subsection{Training protocol} \label{sec:DiscrModels}
To have a reference of the performance of the metrics, different biased and unbiased Face Recognition models have been trained using the base model explained in the Section \ref{sec:Models}.

Basically, a finetuning was carried out: a dense layer was added at the end of the model and trained with different data depending on the aimed bias. For example, to positively bias Asian ethnicity, the dense layer is trained only with faces of Asian people. If non-bias is intended, this dense layer is trained with data from all ethnicities. Triplet Loss function \cite{schroff2015facenet,2022_PR_SetMargin_Morales} is used, whose objective is to bring the feature embeddings of the same user closer together and to pull apart those of different users, in the feature space.

For this training, $50\%$ of the RFW database users were used. To avoid that the results depend on the training of a single model, $M$ models are trained ($M=20$ in this work) for each aimed bias. In this way, we obtain average results not affected by the stochastics associated with the training process. The $M$ models are trained by bootstrapping the $75\%$ of the users within the $50\%$ belonging to the training set. Bootstrapping \cite{efron1992introduction} is a method of inferring results of a population from the results found in a collection of smaller random samples from that population, using replacement during the sampling process.

\subsection{Evaluation protocol}
The other $50\%$ of the RFW users serve for the evaluation stage. In this stage, the values of the $4$ metrics explained in Section \ref{sec:DistrMethods} (SP, EOP, T-Test, N-Sigma) were obtained for each model. These metric values were obtained by ethnic group to compare performance differences between ethnic groups and to establish the existence or not of bias. 

For clarification, a model was trained to be positively biased for one ethnicity and then tested on all ethnicities independently in order to study the differences in performance across them.

Measuring bias in a model is not trivial, it can yield results that are difficult to interpret and compare. When a performance value (for example, EER or TPR) is obtained from a neural network, it has a margin of error, i.e., that value may be a little higher or a little lower depending on the specific data used. To try to avoid this variability and not lose valuable information, in this work the metrics for measuring bias, are obtained from a group of $K$ EER/TPR values (each value obtained from a subset). In other words, several EER/TPR values are calculated to reflect the variability of the performance results in a model and in \textbf{1) Pointwise metrics (SP, EOP)}: performance values are compared one by one, or in \textbf{2) distribution metrics (T-Test, N-Sigma)}: the group of performance values is compared as a distribution.

\subsection{Optimizing subsets} \label{sec:EvalProt} 
 For the computation of the $4$ metrics analyzed in this work, $K$ subsets are created within each ethnic group, and each of these subsets is used to obtain one EER/TPR value. The $K$ subsets have to be representative of the database to obtain a valid set of performance values. Bootstrapping \cite{efron1992introduction} as explained before is a statistical technique that enables us to estimate the characteristics of a population by taking multiple random samples from it. The method involves creating smaller subsets from the larger population, with replacement, and using them to calculate the desired statistics:

\begin{enumerate}
    \item A small number of samples in a subset will give values dependent on the samples selected.
    
    \item A large number of samples may be unnecessary and may complicate the computation. 
\end{enumerate}


\subsection{Experimental framework}

The entire workflow is presented in Figure \ref{fig:Experimental_Framework}:
\begin{enumerate}
    \item The RFW database is divided in training ($50\%$) and evaluation ($50\%$). 
    \item The training stage is used to create the Biased Models as explained in the Subsection \ref{sec:DiscrModels}. As a result, we have $M$ biased models for each of the 4 ethnic groups and $M$ unbiased models, in total: $M\times4 + M$ models.
    \item At the evaluation stage, $K$ subsets of each ethnic group (A,B,C,D) are selected from the database, that is, $K\times4$. Consequently, the models are evaluated with $K\times4$ randomly selected subsets. $K\times4$ new subsets are sampled for each batch of models. As $M$ models of each type are trained ($M$ batches of models), $(K\times4)\times(M)$ subsets are created.
    \item Each evaluation metric (SP, EOP, T-Test, N-Sigma) produces a single value for each ethnicity. Thus, initially, we obtain  $(\textit{number of ethnicities}=4)\times M$ values on each evaluation metric for each of the model types (ethnically biased A,B,C,D and unbiased U). 
    \item $M$ models per type were trained to avoid stochastic effects associated with training in the results. Therefore, the results presented in Section (\ref{sec:results}) show the average over the $M$ model batches. For this reason, the final results show $\textit{number of ethnicities}\times 1$ values on each evaluation metric for each type of model (A,B,C,D,U). 
\end{enumerate}

\section{Measuring Bias in AI Applications}\label{sec:DistrMethods}

Bias refers to the unequal behavior of an algorithm; this irregular behavior may render its decisions unfair and is therefore called biased. Thus, in AI terms, bias is measured in terms of differences in performance between different groups.


In this work we are going to focus on measuring bias using the experimental protocol previously described applied on two pointwise metrics and two distribution metrics.

\subsection{Pointwise metrics}
The experimental protocol is applied to two metrics used in the literature and particularized to our use case. Therefore, the definitions of the metrics undergo certain changes which are shown below. 

Consider a binary classifier $\hat{Y}$. An outcome $\hat{Y}=0$ represents a “non-match” decision (i.e., comparison between samples of different classes), while $\hat{Y}=1$ represents a “match” decision (i.e., comparison between samples from the same class). The literature includes specific measures proposed to detect biased results in machine learning models \cite{barocas2017fairness}. In this work we will use two: 

\begin{itemize}
    \item Statistical Parity (SP) or Demographic Parity \cite{barocas2017fairness}: $P(\bm{\hat{Y}} | \bm{s}=0) = P(\bm{\hat{Y}} | \bm{s}=1)$ which means that the predictions must be independent of attribute $s$, and the probability of obtaining an outcome must always be the same regardless of the attribute (e.g. gender, ethnicity, age). This metric is not suitable for FR as we already explained in Section 2, however in this case we are going to use a particularity of this definition which can be adequate to measure the bias in FR systems. Statistical parity between groups can be expressed in terms of both False Match Rate (FMR) and False Non-Match Rate (FNMR) at a certain decision threshold \cite{marcel2021fairness}:
    \begin{equation}
        \label{eqn:sp_completa}
        \mathrm{SP(\tau)} = 1 - (\alpha A(\tau) + (1 - \alpha)B(\tau))
    \end{equation}
    
    where $\alpha$ defines the weight of the importance of False Matches, $A(\tau)$ is an specific operational point (defined by the threshold $\tau$) of the FMR differential across groups for a given threshold and $B(\tau)$ is the FNMR for this operational point. If taken the threshold at which FMR and FNMR are equal, then the FMR and FNMR became the EER (Equal Error Rate) and $A = B$. The equation simplifies to:
    
    \begin{equation}
        \label{eqn:sp_simplified}
        \mathrm{SP(\tau)} = 1 - A(\tau_{EER}) = 1 - B(\tau_{EER})
    \end{equation}

    Defining $A()$ as the mean of the EER differences between two groups ($G1$ and $G2$), where each EER is obtained from each of the $K$ subsets of the group, we get:
    
    \begin{equation}
        \label{eqn:SP}
        \mathrm{SP(\tau_{EER})=1-\frac{1}{\textit{K}}\sum^{\textit{K}}_{\textit{i}=1}|EER_\textit{i}^{\textit{G}1} - EER_\textit{i}^{\textit{G}2}|}
    \end{equation}

    \item Equality of Opportunity (EOP) \cite{ hardt2016equality}: $P(\bm{\hat{Y}}=1 | \bm{s}=0, \bm{Y}=1 ) = P(\bm{\hat{Y}}=1 | \bm{s}=1, \bm{Y}=1)$. Used in biometric literature as \textit{differential} value, this metric is a relaxed version of the equalized odds criterion. EOP considers only the True Positive Rates (TPR). This definition of EOP serves to indicate that the TPR between different groups must be equal. In this case, the formula depends on the operational point (threshold $\tau$): 
    \begin{equation}
        \label{eqn:EOP}
        \mathrm{EOP(\tau)=1-\frac{1}{\textit{K}}\sum^{\textit{K}}_{\textit{i}=1}|TPR_\textit{i}^{\textit{G}1}(\tau) - TPR_\textit{i}^{\textit{G}2}(\tau)|}
    \end{equation}

The operational point chosen in this work is the one corresponding to an FPR of 0.01.
    
\end{itemize}

\subsection{Distribution metrics}

Instead of comparing the performance values (TPR/EER) independently as it is done with the previous metrics (Equations \ref{eqn:SP} and \ref{eqn:EOP}), in the distribution metrics the performance values are understood as a group, and the aim is to compare them as a distribution.

\subsubsection{Traditional statistical test (T-Test)} 




In our case, an appropriate statistical approach to compare the distributions is the T-Test. This test is used to evaluate the statistical significance of the difference between the means $\mu_{G1}$ and $\mu_{G2}$ of two populations, in situations where the populations follow a normal distribution, the standard deviation is unknown, and the sample size is small. It uses an estimation of the standard deviation instead of the true value.
The selected statistic in our case is the \textit{Welch corrected unpaired T-Test}:

\begin{equation}
    \label{eqn:T-Test}
    Z = \frac{\mu_{G1}-\mu_{G2}}{s} \quad \textrm{where} \quad
    s = \sqrt(\frac{s^2_{G1}+s^2_{G2}}{n})
\end{equation}

\noindent where $n = n_{G1} = n_{G2}$ is the number of samples and $s^2_{G1}$ and $s^2_{G2}$ are the unbiased estimators of the population variance. The null hypothesis $H_0$ : $G1$ = $G2$ (the distribution of FR results for both groups is the same) is rejected if $|Z| >
t_{1 - \alpha/2}$, where $t_{\gamma}$ is the $\gamma$-quantile value of the $t$ distribution.

\subsubsection{The N-Sigma method}



The 5-Sigma method in particle physics refers to the probability in a mass spectrum of having a statistical fluctuation (a peak) in the background. The probability ($p$-value) of a chance peak must not exceed 5$\sigma$ of a normalised Gaussian distribution. The sigma ($\sigma$) is the deviation from the mean ($\mu$) of the distribution that includes approximately 68\% (34\% on each side of the mean) of the data. If we select two sigmas from the mean, we would have around the 95\% of the data. 
If we select 5 sigma, the samples not included are only about a $3 \times 10^{-7}$\%.

When searching for discovery, the data statistic used to discriminate between background only (known as the null hypothesis $H_{0}$) and ``background plus signal" ($H_{1}$) is usually the $L_{1}$/$L_{0}$ likelihood ratio for the two hypotheses; and the 5$\sigma$ criterion is applied to the observed value of this ratio, as compared with its expected distribution assuming just background \cite{lyons2013sigma}.
In this work the N-Sigma method can be expressed as: 

\begin{equation}
    \label{eqn:Nsigma}
    N = \frac{\mu_{G1}-\mu_{G2}}{\sigma_{G1}}
\end{equation}

\noindent where $\mu_{G1}$ and $\mu_{G2}$ are the means of the two populations being compared. $\sigma_{G1}$ is the standard deviation of the population used as reference.

Here, unlike in the T-Test, we do not reject or accept the hypothesis by setting a threshold. In this case the result yields a distance $N$ between the two distributions. This distance can be used to define risk levels.




\section{Experiments}





\subsection{Results} \label{sec:results}

First of all, in Table \ref{Table:EERMean} we have the mean EER/TPR value (\%) for each ethnicity group on the models created. Also, the values of the evaluation metrics are shown in Table \ref{Table:PerfValues}. It is important to understand that the metrics in our work are applied with an evaluating group with respect to a reference group, since bias is a human concept which must be measured with respect to something. The reference group (named as $G1$ in Equation \ref{eqn:SP}, \ref{eqn:EOP}, \ref{eqn:T-Test} and \ref{eqn:Nsigma}) has been chosen as the one with the lowest mean in Table \ref{Table:EERMean} and a similarity value is given for all evaluation groups ($G2$ in Equation \ref{eqn:SP}, \ref{eqn:EOP}, \ref{eqn:T-Test} and \ref{eqn:Nsigma}) with respect to it. The reference group could very well be another one (e.g. the one with the highest mean). In the case of the EOP, SP, and the T-Test, a higher value implies more similarity, although understanding the meaning of the value is not trivial. In the case of the N-Sigma method, a lower value implies more similarity, the specific value being the distance between the distributions in sigmas. In the case that the reference group ($G1$) is the same as the group being evaluated ($G2$) the SP, EOP and T-Test metrics will give the maximum value which is 1, while the N-Sigma metric the minimum which is 0. This is because the groups are identical.

All the values present in Tables \ref{Table:EERMean} and \ref{Table:PerfValues} are an average of the values obtained for the $M=20$ trained models.

\begin{table}[]
    \centering
    \resizebox{\columnwidth}{!}{%
    \begin{tabular}{l|cccc|}      
    \cline{2-5}   
    \multicolumn{1}{l|}{} & \multicolumn{4}{c|}{Ethnicity Evaluated} \\ \hline
    \multicolumn{1}{|l|}{Eth. Finetuned} & \multicolumn{1}{c|}{African} & \multicolumn{1}{c|}{Asian} & \multicolumn{1}{c|}{Caucasian} & Indian \\ \hline
    \multicolumn{1}{|l|}{All} & \multicolumn{1}{c|}{1.83/78.8} & \multicolumn{1}{c|}{1.58/78.1} & \multicolumn{1}{c|}{1.69/82.1} & 2.07/79.1 \\ \hline
    \multicolumn{1}{|l|}{African} & \multicolumn{1}{c|}{1.98/78.3} & \multicolumn{1}{c|}{1.88/74.4} & \multicolumn{1}{c|}{2.04/78.4} & 2.31/74.3 \\ \hline
    \multicolumn{1}{|l|}{Asian} & \multicolumn{1}{c|}{2.26/73.4} & \multicolumn{1}{c|}{1.72/76.6} & \multicolumn{1}{c|}{1.95/76.7} & 2.51/72.3 \\ \hline
    \multicolumn{1}{|l|}{Caucas.} & \multicolumn{1}{c|}{2.15/74.6} & \multicolumn{1}{c|}{1.82/73.7} & \multicolumn{1}{c|}{1.95/79.3} & 2.33/73.3 \\ \hline
    \multicolumn{1}{|l|}{Indian} & \multicolumn{1}{c|}{2.19/73.6} & \multicolumn{1}{c|}{1.88/72.7} & \multicolumn{1}{c|}{2.07/77.1} & \multicolumn{1}{c|}{2.22/76.7} \\ \hline
    \end{tabular}}
    \caption{EER/TPR mean values (\%) for each ethnicity after unbiased (none, first row) and biased training (performing a fine tuning for an specific ethnicity, following 4 rows).}
    \label{Table:EERMean}
\end{table}

\begin{table}
  \centering
  \setlength\tabcolsep{4pt}
  \caption{{The next five subtables (a-e) present performance metrics Mean for the models favored in different ethnicities.}}
  \label{Table:PerfValues}
  \subfloat[Finetuned for All Ethnicities (U). Sigma value: 0.196 ]{%
    \label{Table:PerfValuesAll}%
    \begin{tabular}{|l|c|c|c|c|}
    \hline
    \multicolumn{1}{|l|}{Eth Eval} & \multicolumn{1}{c|}{T-Test} & \multicolumn{1}{c|}{EOP} & \multicolumn{1}{c|}{SP} & \multicolumn{1}{c|}{N-sig} \\ \hline
    African & 1.00$\times10^{-09}$ & 0.967 & 0.997 & 1.30 \\ \hline
    Asian & 1.00 & 0.961 & 1.00 & 0.00 \\ \hline
    Caucasian & 0.16 & 1.00 & 0.998 & 0.58 \\ \hline
    Indian & 4.94$\times10^{-23}$ & 0.970 & 0.995 & 2.53 \\ \hline
    \end{tabular}}

  \vspace{4pt}
  
  \subfloat[Finetuned for African (A). Sigma value: 0.206 ]{%
    \label{Table:PerfValuesAfrican}
    \begin{tabular}{|l|c|c|c|c|}
    \hline
    \multicolumn{1}{|l|}{Eth Eval} & \multicolumn{1}{c|}{T-Test} & \multicolumn{1}{c|}{EOP} & \multicolumn{1}{c|}{SP} & \multicolumn{1}{c|}{N-sig} \\ \hline
    African & 0.12 & 0.998 & 0.998 & 0.52 \\ \hline
    Asian & 1.00 & 0.959 & 1.00 & 0.00 \\ \hline
    Caucasian & 1.00$\times10^{-03}$ & 1.00 & 0.998 & 0.81 \\ \hline
    Indian & 2.81$\times10^{-16}$ & 0.959 & 0.995 & 2.10 \\ \hline
    \end{tabular}}%
  \hfill%
  \subfloat[Finetuned for Asian (B). Sigma value: 0.205]{%
    \label{Table:PerfValuesAsian}
    \begin{tabular}{|l|c|c|c|c|}
    \hline
    \multicolumn{1}{|l|}{Eth Eval} & \multicolumn{1}{c|}{T-Test} & \multicolumn{1}{c|}{EOP} & \multicolumn{1}{c|}{SP} & \multicolumn{1}{c|}{N-sig} \\ \hline
    African & 8.96$\times10^{-31}$ & 0.966 & 0.994 & 2.64 \\ \hline
    Asian & 1.00 & 0.999 & 1.00 & 0.00 \\ \hline
    Caucasian & 1.62$\times10^{-07}$ & 1.00 & 0.997 & 1.16 \\ \hline
    Indian & 5.41$\times10^{-51}$ & 0.955 & 0.992 & 3.89 \\ \hline
    \end{tabular}}%

  \vspace{4pt}
  
  \subfloat[Finetuned for Caucasian (C). Sigma value: 0.210]{%
    \label{Table:PerfValuesCaucasian}
    \begin{tabular}{|l|c|c|c|c|}
    \hline
    \multicolumn{1}{|l|}{Eth Eval} & \multicolumn{1}{c|}{T-Test} & \multicolumn{1}{c|}{EOP} & \multicolumn{1}{c|}{SP} & \multicolumn{1}{c|}{N-sig} \\ \hline
    African & 1.44$\times10^{-14}$ & 0.952 & 0.996 & 1.58 \\ \hline
    Asian & 1.00 & 0.943 & 1.00 & 0.00 \\ \hline
    Caucasian & 6.0$\times10^{-3}$ & 1.00 & 0.998 & 0.59 \\ \hline
    Indian & 1.70$\times10^{-24}$ & 0.938 & 0.994 & 2.32 \\ \hline
    \end{tabular}}%
  \hfill%
  \subfloat[Finetuned for Indian (D). Sigma value: 0.207]{%
    \label{Table:PerfValuesIndian}
    \begin{tabular}{|l|c|c|c|c|}
    \hline
    \multicolumn{1}{|l|}{Eth Eval} & \multicolumn{1}{c|}{T-Test} & \multicolumn{1}{c|}{EOP} & \multicolumn{1}{c|}{SP} & \multicolumn{1}{c|}{N-sig} \\ \hline
    African & 5.16$\times10^{-18}$ & 0.964 & 0.996 & 1.56 \\ \hline
    Asian & 1.00 & 0.954 & 1.00 & 0.00 \\ \hline
    Caucasian & 2.0$\times10^{-4}$ & 1.00 & 0.998 & 0.94 \\ \hline
    Indian & 3.60$\times10^{-27}$ & 0.995 & 0.996 & 1.62 \\ \hline
    \end{tabular}}%
\end{table}

\subsubsection{Discussion}

The best results in terms of mean are achieved with the model trained with all the ethnicities (Table \ref{Table:EERMean}) because they self-regulate each other. In this case, it can be seen that for the EER, the best results are achieved with the Asian ethnicity, while for the TPR the best results are obtained with the Caucasian ethnicity. 

Having seen this, analyzing the tables it is observed that when the model is biased for a particular ethnicity, the similarity value increases with respect to that ethnicity for all the metrics (these results can be seen in subtables \ref{Table:PerfValuesAfrican},\ref{Table:PerfValuesAsian},\ref{Table:PerfValuesCaucasian},\ref{Table:PerfValuesIndian} when compared to subtable \ref{Table:PerfValuesAll}). The Caucasian ethnicity is the exception: training only with the Caucasian samples has not decreased the bias with respect to this ethnicity. We do not have a groundtruth that tells us what should come out and therefore we cannot say in terms of values which metric works best, but must speak in other terms such as interpretability. 

Regarding the interpretability of the different metrics, different aspects can be analyzed.
\begin{itemize}
    \item The EOP and the SP simply measure differences at the subset level and gives you an average value. So what you can see here is a result based simply on a mean difference in performance between groups, and the results should be understood as such.
    \item The interpretability of the T-Test and the N-Sigma is a bit different. In this case, you take the values by subsets and make a comparison of the distributions between the ethnic groups. Therefore, the value tells how far apart the distributions of values are. Both metrics give a value of the difference between the distributions, however, the results of the T-Test are somewhat more difficult to interpret. In this case, a significance level that can be alpha = 0.05 is chosen, and whenever the T-Test result is below that level, it is said that the distributions are statistically different. Subsequently, the lower the value, the more different they are, but the values themselves have no meaning. As for the N-Sigma method, the results represent the same thing but it is easier to interpret what the value itself means. What it means is, with respect to the variance of the favored distribution, how far away the rest of the distributions are, i.e., a value of 1 means that it is at 1 variance and a value of 2 would imply that it is twice as far away.
\end{itemize}

Dissecting the results we realize that with the distribution comparison methods the difference between the ethnic groups can be appreciated more clearly and allows deeper analysis. And between these two methods, the N-Sigma method offers more easily understandable results. 

As we have just explained, N-Sigma and T-Test values do not represent the same thing, one is the result of a statistical test while the other is a distance between distributions. Therefore, although the T-Test and N-Sigma results show some correlation (i.e. lower T-Test values are related to higher N-Sigma values), it is not a perfect inverse correlation (otherwise both values would mean the same thing without contributing anything new). For this reason, for a nearly same value of N-Sigma (0.58 in Table \ref{Table:PerfValuesAll} Eth eval = Caucasian and 0.59 in Table \ref{Table:PerfValuesCaucasian} Eth eval = Caucasian) the T-Test value is different.

\section{Conclusion}

In this paper we propose the use of metrics to measure bias under an experimental protocol and specifically, a metric called N-Sigma widely used in other fields but unexplored in AI. This metric is based on the well-established idea of \textit{5-sigmas} used in fields such as physics or economics. We have evaluated a model fine-tuned to be biased for different ethnicities with the distribution and pointwise metrics.

The results show that the distribution methods yield results more interpretable. Among the distribution comparison methods the N-Sigma method results are more user-friendly. The use of this metric makes it possible to adapt very easily to different use cases by varying the sigma at which bias is considered to occur, e.g., defining different risk levels associated with different values of N. In applications where the presence of bias is critical (high risk), a lower sigma value can be assigned, while if the application is more flexible in this regard, the allowed sigma value can be increased.


\section*{Acknowledgment}
Support by project BBforTAI (PID2021-127641OB-I00 MICINN/FEDER). D. deAlcala is supported by a FPU Fellowship (FPU21/05785) from the Spanish MIU.

\bibliographystyle{ieeetr}
\bibliography{egbib.bib}

\end{document}